\documentclass[11pt,a4paper]{article}
\usepackage[hyperref]{acl2020}
\usepackage[utf8]{inputenc}
\usepackage{times}
\usepackage{latexsym}

\usepackage{microtype}
\usepackage{amsfonts}
\usepackage{eucal}
\usepackage{bm}
\usepackage{graphicx}
\usepackage{adjustbox}

\usepackage{booktabs}
\usepackage{siunitx}

\usepackage{todonotes}

\aclfinalcopy 

\title{On Exposure Bias, Hallucination and Domain Shift\\ in Neural Machine Translation}

\author{Chaojun Wang$^1$ \quad Rico Sennrich$^{2,1}$ \bigskip\\
  $^1$School of Informatics, University of Edinburgh \\
  $^2$Department of Computational Linguistics, University of Zurich \\
  \texttt{zippo\_wang@foxmail.com, sennrich@cl.uzh.ch}
  }

\date{}

\begin{document}
\maketitle
\begin{abstract}
The standard training algorithm in neural machine translation (NMT) suffers from exposure bias, and alternative algorithms have been proposed to mitigate this.
However, the practical impact of exposure bias is under debate.
In this paper, we link exposure bias to another well-known problem in NMT, namely the tendency to generate hallucinations under domain shift.
In experiments on three datasets with multiple test domains, we show that exposure bias is partially to blame for hallucinations, and that training with Minimum Risk Training, which avoids exposure bias, can mitigate this.
Our analysis explains why exposure bias is more problematic under domain shift, and also links exposure bias to the beam search problem, i.e.\ performance deterioration with increasing beam size.
Our results provide a new justification for methods that reduce exposure bias: even if they do not increase performance on in-domain test sets, they can increase model robustness to domain shift.
\end{abstract}

\section{Introduction}
Neural Machine Translation (NMT) has advanced the state of the art in MT~\citep{NIPS2014_5346, DBLP:journals/corr/BahdanauCB14, NIPS2017_7181}, but is susceptible to domain shift.
\citet{koehn-knowles-2017-six} consider out-of-domain translation one of the key challenges in NMT.
Such translations may be fluent, but completely unrelated to the input (\textit{hallucinations}), and their misleading nature makes them particularly problematic.

We hypothesise that \textit{exposure bias}~\citep{DBLP:journals/corr/RanzatoCAZ15}, a discrepancy between training and inference, makes this problem worse.
Specifically, training with teacher forcing only exposes the model to gold history, while previous predictions during inference may be erroneous.
Thus, the model trained with teacher forcing may over-rely on previously predicted words, which would exacerbate error propagation.
Previous work has sought to reduce exposure bias in training~\citep{NIPS2015_5956, DBLP:journals/corr/RanzatoCAZ15, shen-etal-2016-minimum, wiseman-rush-2016-sequence, zhang-etal-2019-bridging}.
However, the relevance of error propagation is under debate: \citet{wu-etal-2018-beyond} argue that its role is overstated in literature, and that linguistic features explain some of the accuracy drop at higher time steps.

Previous work has established a link between domain shift and hallucination in NMT \citep{koehn-knowles-2017-six,mller2019domain}.
In this paper, we will aim to also establish an empirical link between hallucination and exposure bias.
Such a link will deepen our understanding of the hallucination problem, but also has practical relevance, e.g.\ to help predicting in which settings the use of sequence-level objectives is likely to be helpful.
We further empirically confirm the link between exposure bias and the `beam search problem', i.e.\ the fact that translation quality does not increase consistently with beam size~\citep{koehn-knowles-2017-six, ott2018analyzing, stahlberg-byrne-2019-nmt}.

We base our experiments on German$\to$English IWSLT'14, and two datasets used to investigate domain robustness by \citet{mller2019domain}: a selection of corpora from OPUS~\citep{lison-tiedemann-2016-opensubtitles2016} for German$\to$English, and a low-resource German$\to$Romansh scenario.
We experiment with Minimum Risk Training (MRT)~\citep{och-2003-minimum, shen-etal-2016-minimum}, a training objective which inherently avoids exposure bias.

Our experiments show that MRT indeed improves quality more in out-of-domain settings, and reduces the amount of hallucination.
Our analysis of translation uncertainty also shows how the MLE baseline over-estimates the probability of random translations at all but the initial time steps, and how MRT mitigates this problem.
Finally, we show that the beam search problem is reduced by MRT.

\section{Minimum Risk Training}

The de-facto standard training objective in NMT is to minimize the negative log-likelihood $\mathcal{L}(\boldsymbol{\theta})$ of the training data $D$\footnote{This is equivalent to maximizing the likelihood of the data, hence \textit{Maximum Likelihood Estimation} (MLE).}:
\begin{equation}
\label{eq:1}
\mathcal{L}(\boldsymbol{\theta})=\sum_{(\mathbf{x},\mathbf{y}) \in D} \sum_{t=1}^{|\mathbf{y}|} -\log P\left(\mathbf{y}_{t} | \mathbf{x}, \mathbf{y}_{<t} ; \boldsymbol{\theta}\right)
\end{equation}
where $\mathbf{x}$ and $\mathbf{y}$ are the source and target sequence, respectively, $\mathbf{y}_t$ is the t\textsuperscript{th} token in $\mathbf{y}$, and $\mathbf{y}_{<t}$ denotes all previous tokens. MLE is typically performed with teacher forcing, where $\mathbf{y}_{<t}$ are ground-truth labels in training, which creates a mismatch to inference, where $\mathbf{y}_{<t}$ are model predictions.

Minimum Risk Training (MRT) is a sequence-level objective that avoids this problem. Specifically, the objective function of MRT is the expected loss (\textit{risk}) with respect to the posterior distribution:
\begin{equation}
\label{eq:2}
\mathcal{R}(\boldsymbol{\theta})=\sum_{(\mathbf{x},\mathbf{y}) \in D} \sum_{\mathbf{\tilde{y}} \in \mathcal{Y}(\mathbf{x})}P\left(\mathbf{\tilde{y}} | \mathbf{x} ; \boldsymbol{\theta}\right) \Delta\left(\mathbf{\tilde{y}}, \mathbf{y}\right)
\end{equation}
in which the loss $\Delta\left(\mathbf{\tilde{y}}, \mathbf{y}\right)$ indicates the discrepancy between the gold translation $\mathbf{y}$ and the model prediction $\mathbf{\tilde{y}}$. Due to the intractable search space, the posterior distribution $\mathcal{Y}(\mathbf{x})$ is approximated by a subspace $\mathcal{S}(\mathbf{x})$ by sampling a certain number of candidate translations, and normalizing:
\begin{equation}
\label{eq:3}
\tilde{P}\left(\mathbf{\tilde{y}} | \mathbf{x} ; \boldsymbol{\theta}, \alpha\right)=\frac{P\left(\mathbf{\tilde{y}} | \mathbf{x} ; \boldsymbol{\theta}\right)^{\alpha}}{\sum_{\mathbf{y}^{\prime} \in \mathcal{S}\left(\mathbf{x}\right)} P\left(\mathbf{y}^{\prime} | \mathbf{x} ; \boldsymbol{\theta}\right)^{\alpha}}
\end{equation}
where $\alpha$ is a hyperparameter to control the sharpness of the subspace.
Based on preliminary results, we use random sampling to generate candidate translations, and following~\citet{edunov-etal-2018-classical}, do not add the reference translation to the subspace.

\section{Experiments}
\subsection{Data}
To verify the effectiveness of our MRT implementation on top of a strong Transformer baseline \citep{NIPS2017_7181}, we first conduct experiments on IWSLT'14 German$\to$English (DE$\to$EN) \citep{cettolo2014report}, which consists of \num{180000} sentence pairs. We follow previous work for data splits~\citep{DBLP:journals/corr/RanzatoCAZ15,edunov-etal-2018-classical}.

For experiments with domain shift, we use data sets and preprocessing as~\citet{mller2019domain}\footnote{\url{https://github.com/ZurichNLP/domain-robustness}}.
For DE$\to$EN, data comes from OPUS~\citep{lison-tiedemann-2016-opensubtitles2016}, and is comprised of five domains: \textit{medical}, \textit{IT}, \textit{law}, \textit{koran} and \textit{subtitles}.
We use \textit{medical} for training and development, and report results on an in-domain test set and the four other domains (out-of-domain; OOD).
German$\to$Romansh (DE$\to$RM) is a low-resource language pair where robustness to domain shift is of practical relevance. The training data is from the Allegra corpus~\citep{scherrer-cartoni-2012-trilingual} (\textit{law} domain) with \num{100000} sentence pairs. The test domain are \textit{blogs}, using data from Convivenza\footnote{\url{https://www.suedostschweiz.ch/blogs/convivenza}}.
We have access to 2000 sentences for development and testing, respectively, in each domain.

We tokenise and truecase data sets with Moses~\citep{koehn-etal-2007-moses}, and use shared BPE with \num{32000} units~\citep{sennrich-etal-2016-neural}.

\subsection{Model}
We implement\footnote{Code available at \url{https://github.com/zippotju/Exposure-Bias-Hallucination-Domain-Shift}} MRT in the Nematus toolkit~\citep{sennrich-etal-2017-nematus}. 
All our experiments use the Transformer architecture~\citep{NIPS2017_7181}.
Following ~\citet{edunov-etal-2018-classical}, we use \mbox{$1$-$\text{BLEU}_\text{smooth}$} \citep{lin-och-2004-orange} as the MRT loss. 
Models are pre-trained with the token-level objective MLE and then fine-tuned with MRT.
Hyperparameters mostly follow previous work \citep{edunov-etal-2018-classical,mller2019domain}; for MRT, we conduct limited hyperparameter search on the IWSLT'14 development set, including learning rate, batch size, and the sharpness parameter $\alpha$.
We set the number of candidate translations for MRT to 4 to balance effectiveness and efficiency.
Detailed hyperparameters are reported in the Appendix.

\subsection{Evaluation}

For comparison to previous work, we report lowercased, tokenised BLEU~\citep{papineni-etal-2002-bleu} with \textit{multi-bleu.perl} for IWSLT'14, and cased, detokenised BLEU with SacreBLEU~\citep{post-2018-call}\footnote{Signature: BLEU+c.mixed+\#.1+s.exp+tok.13a+v.1.4.2} otherwise.
For settings with domain shift, we report average and standard deviation of 3 independent training runs to account for optimizer instability.

The manual evaluation was performed by two native speakers of German who completed bilingual (German/English) high school or University programs.
We collected $\sim$3600 annotations in total, spread over 12 configurations.
We ask annotators to evaluate translations according to fluency and adequacy. For fluency, the annotator classifies a translation as fluent, partially fluent or not fluent; for adequacy, as adequate, partially adequate or inadequate. 
We report kappa coefficient ($K$)~\citep{carletta-1996-assessing} for inter-annotator and intra-annotator agreement in Table~\ref{tab:7}, and assess statistical significance with Fisher's exact test (two-tailed).

\begin{table}
\centering
\begin{small}
\setlength{\tabcolsep}{0.5em}
\begin{tabular}{lcccccc}
\toprule
& \multicolumn{3}{c}{\textbf{inter-annotator}} & \multicolumn{3}{c}{\textbf{intra-annotator}} \\
\cmidrule(lr){2-4}\cmidrule(lr){5-7}
annotation & $P(A)$ & $P(E)$ & $K$ & $P(A)$ & $P(E)$ & $K$ \\
\midrule
fluency & 0.66 & 0.38 & 0.44 & 0.87 & 0.42 & 0.77 \\
adequacy & 0.82 & 0.61 & 0.54 & 0.93 & 0.66 & 0.79 \\
\bottomrule
\end{tabular}
\end{small}
\caption{Inter-annotator (N=250) and intra-annotator agreement (N=617) of manual evaluation.}
\label{tab:7}
\end{table}

\subsection{Results}

Table~\ref{tab:1} shows results for IWSLT'14.
We compare to results by \citet{edunov-etal-2018-classical}, who use a convolutional architecture~\citep{Gehring:2017:CSS:3305381.3305510}, and \citet{wu2018pay}, who report results with Transfomer-base and dynamic convolution.
\begin{table}
\centering
\begin{small}
\begin{tabular}{lc}
\toprule
system   & BLEU \\
\midrule
ConvS2S (MLE)~\citep{edunov-etal-2018-classical} & 32.2 \phantom{(+\textbf{0.6})} \\
ConvS2S (MRT)~\citep{edunov-etal-2018-classical} & 32.8 (+\textbf{0.6}) \\
Transformer (MLE)~\citep{wu2018pay} & 34.4 \phantom{(+\textbf{0.6})} \\
DynamicConv (MLE)~\citep{wu2018pay} & 35.2 \phantom{(+\textbf{0.6})} \\
\midrule
MLE  & 34.7 \phantom{(+\textbf{0.6})}  \\
MRT & 35.2 (+\textbf{0.5}) \\
\bottomrule
\end{tabular}
\end{small}
\caption{Results for IWSLT'14 DE$\to$EN with MLE and MRT (in brackets, improvement over MLE).}
\label{tab:1}
\end{table}

\begin{table*}[!t]
\centering
\begin{small}
\begin{tabular}{lcccc}
\toprule
& \multicolumn{2}{c}{\textbf{DE$\to$EN}} & \multicolumn{2}{c}{\textbf{DE$\to$RM}} \\
\cmidrule(lr){2-3}\cmidrule(lr){4-5}
system & in-domain & average OOD & in-domain & average OOD\\
\midrule
SMT~\cite{mller2019domain} & 58.4 \phantom{($\pm$0.00)} & 11.8 \phantom{($\pm$0.00)}& 45.2 \phantom{($\pm$0.00)}& 15.5 \phantom{($\pm$0.00)}\\
NMT~\cite{mller2019domain} & 61.5 \phantom{($\pm$0.00)} & 11.7 \phantom{($\pm$0.00)} & 52.5 \phantom{($\pm$0.00)} & 18.9 \phantom{($\pm$0.00)}\\
NMT+RC+SR+NC~\cite{mller2019domain} & 60.8 \phantom{($\pm$0.00)} & 13.1 \phantom{($\pm$0.00)} & 52.4 \phantom{($\pm$0.00)} & 20.7 \phantom{($\pm$0.00)}\\
\midrule
MLE w/o LS  & 58.3 ($\pm$0.53) & \phantom{0}9.7 ($\pm$0.25)  & 52.2 ($\pm$0.19) & 15.8 ($\pm$0.39)\\
+MRT & 58.4 ($\pm$0.39) & 10.2 ($\pm$0.26) & 52.1 ($\pm$0.08) & 15.9 ($\pm$0.28)\\
MLE w/ LS & 58.9 ($\pm$0.45) & 11.2 ($\pm$0.16) & 53.9 ($\pm$0.16) & 18.0 ($\pm$0.17)\\
+MRT & 58.8 ($\pm$0.36) & 12.0 ($\pm$0.29) & 53.9 ($\pm$0.12) & 18.7 ($\pm$0.09)\\
\bottomrule
\end{tabular}
\end{small}
\caption{Average BLEU and standard deviation on in-domain and out-of-domain test sets for models trained on OPUS (DE$\to$EN) and Allegra (DE$\to$RM). RC: reconstruction; SR: subword regularization, NC: noisy channel.}
\label{tab:2}
\end{table*}

With 34.7 BLEU, our baseline is competitive.
We observe an improvement of 0.5 BLEU from MRT, comparable to \citet{edunov-etal-2018-classical}, although we start from a stronger baseline (+2.5 BLEU).

Table~\ref{tab:2} shows results for data sets with domain shift.
To explore the effect of label smoothing~\citep{7780677}, we train baselines with and without label smoothing.
MLE with label smoothing performs better by itself, and we also found MRT to be more effective on top of the initial model with label smoothing.
For DE$\to$EN, MRT increases average OOD BLEU by 0.8 compared to the MLE baseline with label smoothing; for DE$\to$RM the improvement is 0.7 BLEU.
We note that MRT does not consistently improve in-domain performance, which is a first indicator that exposure bias may be more problematic under domain shift.

Our OOD results lag slightly behind those of \citet{mller2019domain}, but note that the techniques employed by them, namely reconstruction \citep{Tu:2017:NMT:3298483.3298684, niu-etal-2019-bi}, subword regularization \citep{kudo-2018-subword}, and noisy channel modelling \citep{li2016mutual} are orthogonal to MRT. We leave the combination of these approaches to future work.

\section{Analysis}

BLEU results indicate that MRT can improve domain robustness.
In this section, we report on additional experiments to establish more direct links between exposure bias and domain robustness, hallucination, and the beam search problem.
Experiments are performed on DE$\to$EN OPUS data.

\subsection{Hallucination}

We manually evaluate the proportion of hallucinated translations on out-of-domain and in-domain test sets.
We follow the definition and evaluation by ~\citet{mller2019domain}, considering a translation a hallucination if it is \textbf{(partially) fluent}, but unrelated in content to the source text (\textbf{inadequate}). We report the proportion of such hallucinations for each system.

\begin{table}[!ht]
\centering
\small
\begin{tabular}{lcc}
\toprule
& \multicolumn{2}{c}{\% hallucinations (BLEU)} \\
\cmidrule{2-3}
system   & out-of-domain & in-domain \\
\midrule
MLE w/o LS  & 35\% \phantom{0}(9.7) & 2\% (58.3) \\
+MRT & 29\% (10.2) & - \\
MLE w/ LS & 33\% (11.2) & 1\% (58.9)\\
+MRT & 26\% (12.0) & - \\
\bottomrule
\end{tabular}
\caption{Proportion of hallucinations and BLEU on out-of-domain and in-domain test sets. DE$\to$EN OPUS.}
\label{tab:3}
\end{table}

Results in Table~\ref{tab:3} confirm that hallucinations are much more pronounced in out-of-domain test sets (33--35\%) than in in-domain test sets (1--2\%).
MRT reduces the proportion of hallucinations on out-of-domain test sets (N=500 for each system; reductions statistically significant at $p<0.05$) and improves BLEU.
Note that the two metrics do not correlate perfectly: MLE with label smoothing has higher BLEU (+1) than MRT based on MLE without label smoothing, but a similar proportion of hallucinations.
This indicates that label smoothing increases translation quality in other aspects, while MRT has a clear effect on the number of hallucinations, reducing it by up to 21\% (relative).

\begin{table}
\begin{center}
\small
\begin{tabular}{ll}
\toprule
source & \textbf{Wir haben} ihn \textbf{gefunden}. \\
reference & \textbf{We found} him.\\
MLE & Do not pass it. \\
MRT & \textbf{We have found} it.  \\
\midrule
source & So höre \textbf{nicht} auf die \textbf{Ableugner}.\\
reference & So hearken \textbf{not} to those who \textbf{deny} (the Truth).\\
MLE & \textbf{Do not} drive or use machines. \\
MRT & \textbf{Do not} apply to \textbf{dleugner}.  \\
\bottomrule
\end{tabular}
\caption{Out-of-domain translation examples. MLE hallucinates in both examples; MRT was rated more adequate in top example, less fluent in bottom one.}
\label{tab:6}
\end{center}
\end{table}

A closer inspection of segments where the MLE system was found to hallucinate shows that some segments were scored higher in adequacy with MRT, others lower in fluency.
One example for each case is shown in Table~\ref{tab:6}.
Even the example where MRT was considered disfluent and inadequate actually shows an attempt to cover the source sentence: the source word `Ableugner' (denier) is mistranslated into `dleugner'.
We consider this preferable to producing a complete hallucination.

\subsection{Uncertainty Analysis}
\label{sec:4.2}
\begin{figure*}[ht]
    \centering
    \includegraphics[width=15cm]{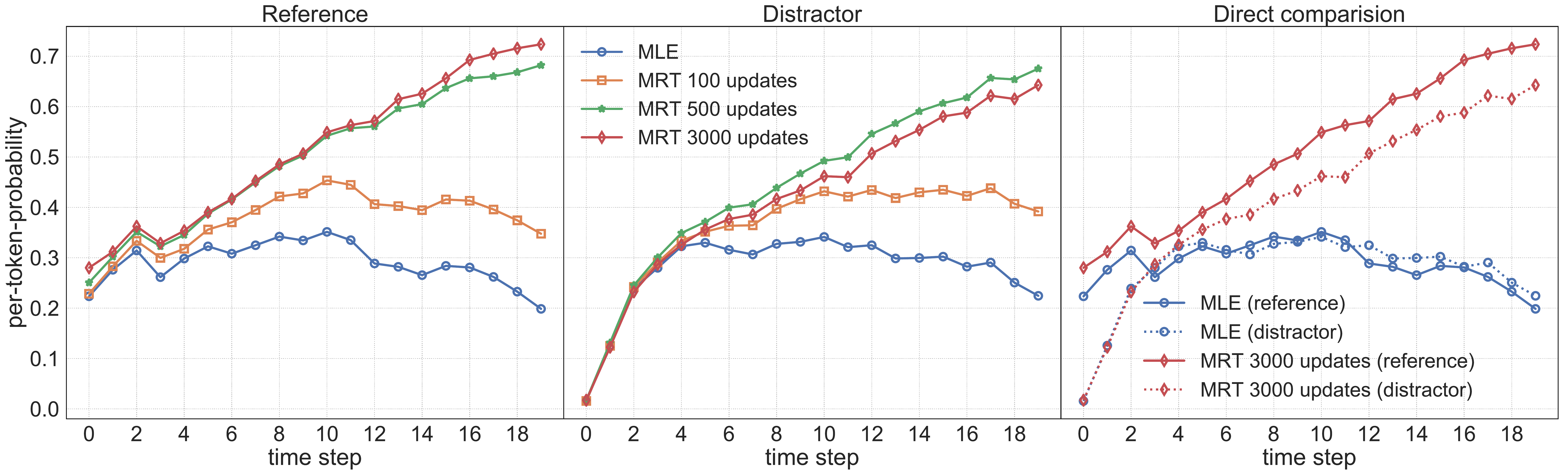}
    \caption{Per-token probability of out-of-domain reference translations and in-domain distractors (first two graphs share legend). Rightmost plot shows direct comparison for MLE baseline and final MRT model. DE$\to$EN OPUS .}
    \label{fig:1}
\end{figure*}

Inspired by~\citet{ott2018analyzing}, we analyse the model's uncertainty by computing the average probability at each time step across a set of sentences.
Besides the reference translations, we also consider a set of `distractor' translations, which are random sentences from the in-domain test set which match the corresponding reference translation in length.

In Figure~\ref{fig:1}, we show out-of-domain results for an MLE model and multiple checkpoints of MRT fine-tuning.
The left two graphs show probabilities for references and distractors, respectively. 
The right-most graph shows a direct comparison of probabilities for references and distractors for the MLE baseline and the final MRT model.
The MLE baseline assigns similar probabilities to tokens in the references and the distractors. 
Only for the first time steps is there a clear preference for the references over the (mostly random!) distractors.
This shows that error propagation is a big risk: should the model make a wrong prediction initially, this is unlikely to be penalised in later time steps.

MRT tends to increase the model's certainty at later time steps\footnote{The uncertainty of the baseline is due to label smoothing.}, but importantly, the increase is sharper for the reference translations than for the distractors.
The direct comparison shows a widening gap in certainty between the reference and distractor sentences.\footnote{For intermediate checkpoints, see Appendix, Figure~\ref{fig:3}.} 
In other words, producing a hallucination will incur a small penalty at each time step (compared to producing the reference), presumably due to a higher reliance on the source signal, lessening the risk of error propagation and hallucinations.

Our analysis shows similar trends on in-domain references.
However, much higher probabilities are assigned to the first few tokens of the references than to the distractors.
Hence, it is much less likely that a hallucination is kept in the beam, or will overtake a good translation in overall probability, reducing the practical impact of the model's over-reliance on its history.\footnote{Figures are shown in the Appendix (Figure~\ref{fig:4}).}

\subsection{Beam Size Analysis}
\label{sec:4.3}

Figure~\ref{fig:1} shows that with MLE, distractor sentences are assigned lower probabilities than the references at the first few time steps, but are assigned similar, potentially even higher probabilities at later time steps.
This establishes a connection between exposure bias and the beam search problem, i.e.\ the problem that increasing the search space can lead to worse model performance.\footnote{The beam search problem has previously been linked to length bias \citep{yang-etal-2018-breaking,murray-chiang-2018-correcting} and the copy mode ~\citep{ott2018analyzing}. We consider hallucinations another result of using large search spaces with MLE models.} With larger beam size, it is more likely that hallucinations survive pruning at the first few time steps, and with high probabilities assigned to them at later time steps, there is a chance that they become the top-scoring translation.

We investigate whether the beam search problem is mitigated by MRT.
In Table~\ref{tab:beam}, we report OOD BLEU and the proportion of hallucinations with beam sizes of 1, 4 and 50.
While MRT does not eliminate the beam search problem, performance drops less steeply as beam size increases.
With beam size 4, our MRT models outperform the MLE baseline by 0.5-0.8 BLEU; with beam size 50, this difference grows to 0.6-1.5 BLEU. 
Our manual evaluation (N=200 for each system for beam size 1 and 50) shows that the proportion of hallucinations increases with beam size, and that MRT consistently reduces the proportion by 11-21\% (relative).
For the system with label smoothing, the relative increase in hallucinations with increasing beam size is also smaller with MRT (+33\%) than with MLE (+44\%).

\begin{table}[ht]
\centering
\small
\begin{tabular}{lccc}
\toprule
& \multicolumn{3}{c}{BLEU (\% hallucinations)} \\
\cmidrule{2-4}
system   & $k=1$ & $k=4$ & $k=50$ \\
\midrule
MLE w/o LS & \phantom{0}8.9 (28\%) & \phantom{0}9.7 (35\%) & \phantom{0}9.3 (37\%)\\
+MRT & \phantom{0}9.1 (24\%) & 10.2 (29\%) & \phantom{0}9.9 (33\%)\\
MLE w/ LS & 10.6 (27\%) & 11.2 (33\%) & \phantom{0}9.4 (39\%)\\
+MRT & 11.3 (24\%) & 12.0 (26\%) & 10.9 (32\%)\\
\bottomrule
\end{tabular}
\caption{Average OOD BLEU and proportion of hallucinations with different beam sizes $k$. DE$\to$EN OPUS.}
\label{tab:beam}
\end{table}

\section{Conclusions}

Our results and analysis show a connection between the exposure bias due to MLE training with teacher forcing and several well-known problems in neural machine translation, namely poor performance under domain shift, hallucinated translations, and deteriorating performance with increasing beam size.
We find that Minimum Risk Training, which does not suffer from exposure bias, can be useful even when it does not increase performance on an in-domain test set: it increases performance under domain shift, reduces the number of hallucinations substantially, and makes beam search with large beams more stable.

Our findings are pertinent to the academic debate how big of a problem exposure bias is in practice -- we find that this can vary substantially depending on the dataset --, and they provide a new justification for sequence-level training objectives that reduce or eliminate exposure bias.
Furthermore, we believe that a better understanding of the links between exposure bias and well-known translation problems will help practitioners decide when sequence-level training objectives are especially promising, for example in settings where the test domain is unknown, or where hallucinations are a common problem.

\section*{Acknowledgments}

Chaojun Wang was supported by the UK Engineering and Physical Sciences Research Council (EPSRC) fellowship grant EP/S001271/1 (MTStretch).
Rico Sennrich acknowledges support of the Swiss National Science Foundation (MUTAMUR; no.\ 176727).
This project has received support from Samsung Electronics Polska sp. z o.o. - Samsung R\&D Institute Poland.

\clearpage
\bibliography{anthology,acl2020}
\bibliographystyle{acl_natbib}

\clearpage
\appendix

\section{Appendix}

\vspace*{\fill}

\noindent\begin{minipage}{\textwidth}
\centering
\begin{tabular}{ccc}
\hline
& IWSLT & OPUS/Allegra \\
\cline{2-3}
\textbf{General hyperparameters}  & &   \\
embedding layer size  & \multicolumn{2}{c}{512} \\
hidden state size & \multicolumn{2}{c}{512} \\
tie encoder decoder embeddings  & \multicolumn{2}{c}{yes} \\
tie decoder embeddings  & \multicolumn{2}{c}{yes} \\
loss function & \multicolumn{2}{c}{per-token-cross-entropy (MRT)} \\
label smoothing & \multicolumn{2}{c}{0.1}\\
optimizer & \multicolumn{2}{c}{adam} \\
learning schedule & \multicolumn{2}{c}{transformer (constant)} \\
warmup steps & 4000 & 6000 \\
gradient clipping threshold & 1 & 0 \\
maximum sequence length &\multicolumn{2}{c}{100}\\
token batch size & \multicolumn{2}{c}{4096}\\
length normalization alpha & 0.6 & 1 \\
encoder depth & \multicolumn{2}{c}{6}\\
decoder depth & \multicolumn{2}{c}{6}\\
feed forward num hidden & 1024 & 2048 \\
number of attention heads & 4 & 8 \\
embedding dropout & 0.3 & 0.1 \\
residual dropout & 0.3 & 0.1 \\
relu dropout & 0.3 & 0.1 \\
attention weights dropout & 0.3 & 0.1 \\
beam size & \multicolumn{2}{c}{4}\\
\hline
& beam search sampling & random sampling \\
\cline{2-3}
\textbf{MRT-revelant hyperparameters} & &\\
learning rate & 0.00003 & 0.00001 \\
batch size & 8192 (tokens) & 10 (sentences) \\
sharpness alpha & 0.005 & 0.005 \\
\hline
\end{tabular}
\captionof{table}{Configurations of NMT systems used to pre-train and fine-tune over three datasets. Note in general hyperparameters, the items in brackets denote the options that will be used in MRT fine-tuning.}
\label{tab:4}
\end{minipage}
\vspace*{\fill}

\newpage

\begin{figure*}[!ht]
    \centering
    \includegraphics[width=16cm]{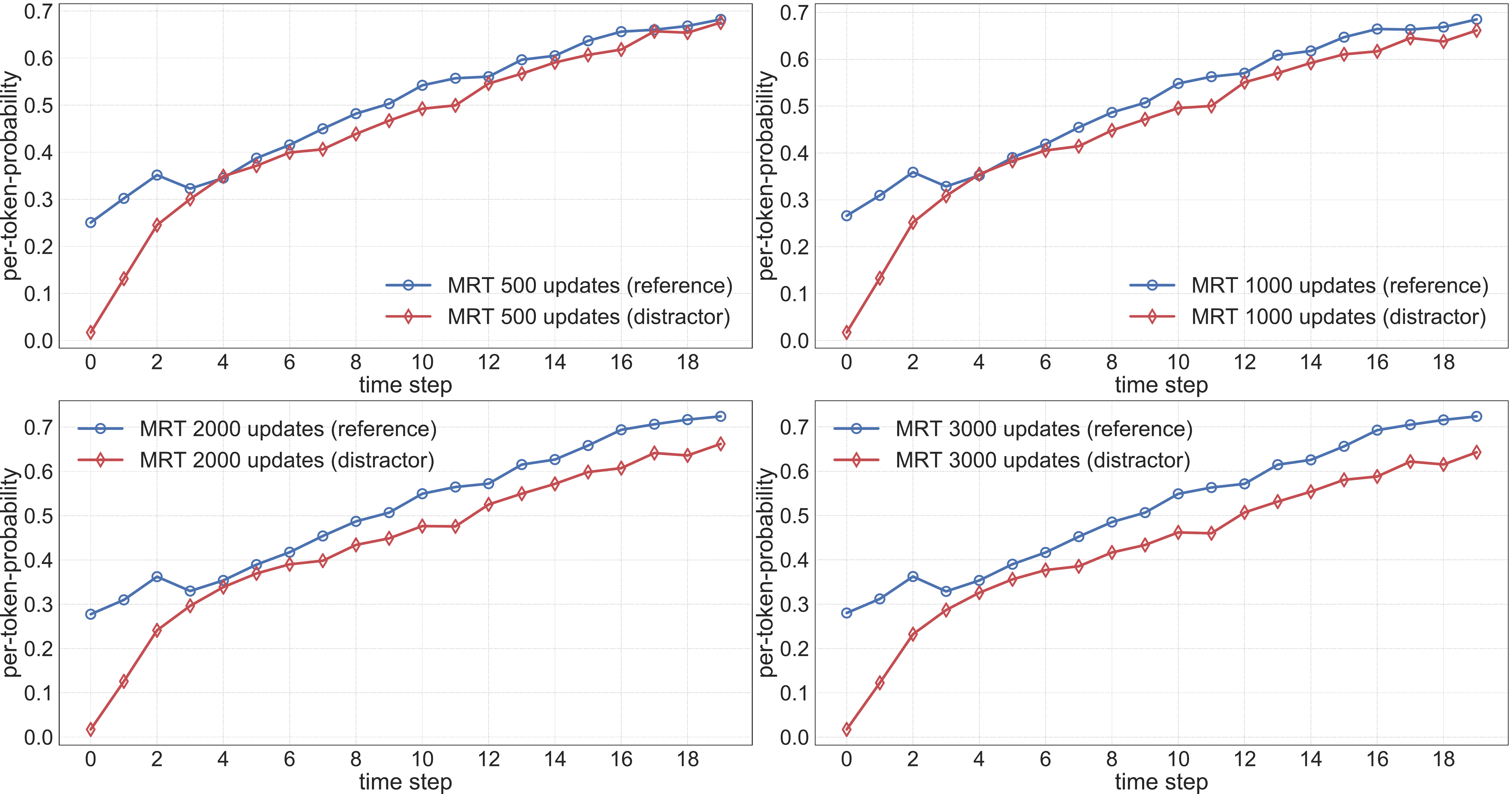}
    \caption{Per-token probability of \textbf{out-of-domain} reference translations and in-domain distractors for different checkpoints in MRT training, showing a widening gap between references and distractors. DE$\to$EN OPUS.}
    \label{fig:3}
\end{figure*}

\begin{figure*}[!ht]
    \centering
    \includegraphics[width=16cm]{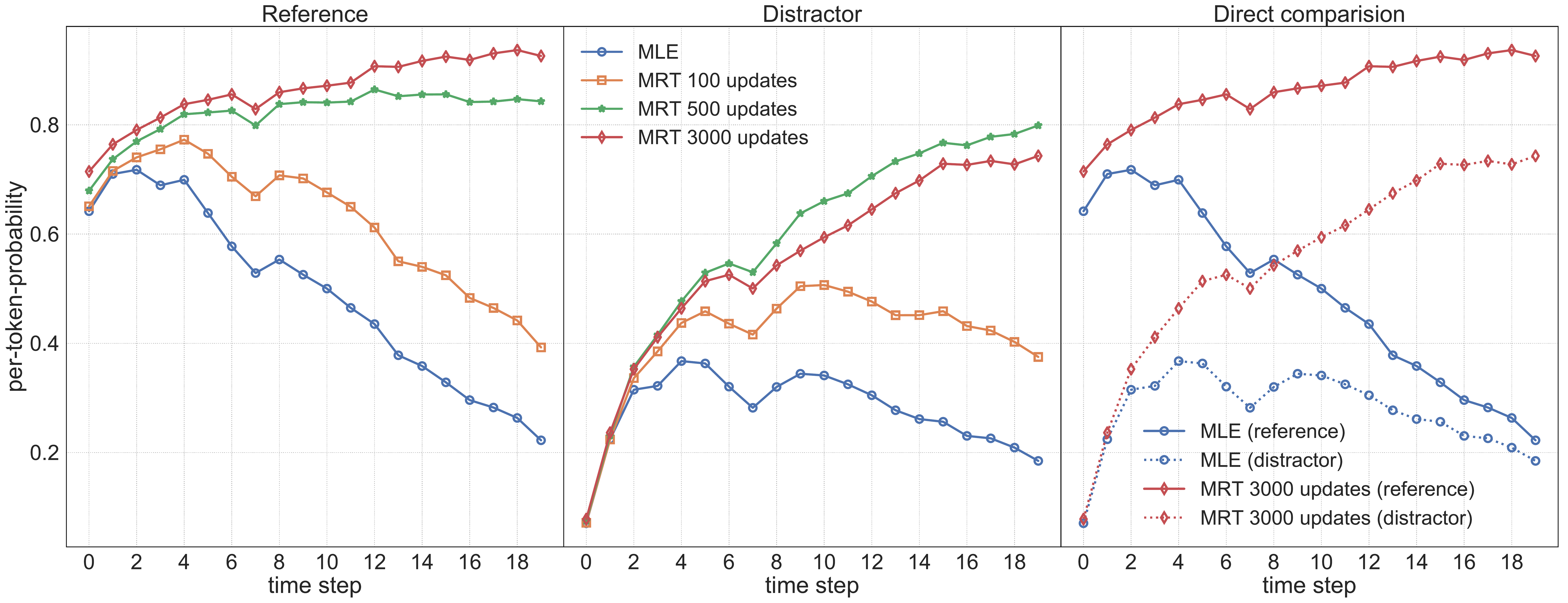}
    \caption{Per-token probability of \textbf{in-domain} reference translations and distractors. Rightmost plot shows direct comparison for MLE baseline and final MRT model. DE$\to$EN OPUS.}
    \label{fig:4}
\end{figure*}

\end{document}